\documentclass[sigconf]{acmart}

\copyrightyear{2020} 
\acmYear{2020} 
\setcopyright{acmcopyright}\acmConference[MM '20]{Proceedings of the 28th ACM International Conference on Multimedia}{October 12--16, 2020}{Seattle, WA, USA}
\acmBooktitle{Proceedings of the 28th ACM International Conference on Multimedia (MM '20), October 12--16, 2020, Seattle, WA, USA}
\acmPrice{15.00}
\acmDOI{10.1145/3394171.3413886}
\acmISBN{978-1-4503-7988-5/20/10}

\usepackage{multirow}
\begin{document}
\fancyhead{}
\title{Topic Adaptation and Prototype Encoding for Few-Shot Visual Storytelling}


\author{Jiacheng Li}
\affiliation{%
  \institution{College of Computer Science and
Technology, ZheJiang University}
}
\email{lijiacheng@zju.edu.cn}

\author{Siliang Tang}
\authornote{Siliang Tang is the corresponding author.}
\affiliation{%
    \institution{College of Computer Science and
Technology, ZheJiang University}
}
\email{siliang@zju.edu.cn}

\author{Juncheng Li}
\affiliation{%
    \institution{College of Computer Science and
Technology, ZheJiang University}
}
\email{junchengli@zju.edu.cn}

\author{Jun Xiao, Fei Wu}
\affiliation{%
    \institution{College of Computer Science and
Technology, ZheJiang University}
}
\email{{junx, wufei}@cs.zju.edu.cn}

\author{Shiliang Pu}
\affiliation{%
  \  \institution{Hikvision Research Institute}
}
\email{pushiliang.hri@hikvision.com}

\author{Yueting Zhuang}
\affiliation{%
    \institution{College of Computer Science and
Technology, ZheJiang University}
}
\email{yzhuang@cs.zju.edu.cn}

\begin{abstract}
Visual Storytelling~(VIST) is a task to tell a narrative story about a certain topic according to the given photo stream. The existing studies focus on designing complex models, which rely on a huge amount of human-annotated data. However, the annotation of VIST is extremely costly and many topics cannot be covered in the training dataset due to the long-tail topic distribution. In this paper, we focus on enhancing the generalization ability of the VIST model by considering the few-shot setting. Inspired by the way humans tell a story, we propose a topic adaptive storyteller to model the ability of inter-topic generalization. In practice, we apply the gradient-based meta-learning algorithm on multi-modal seq2seq models to endow the model the ability to adapt quickly from topic to topic. Besides, We further propose a prototype encoding structure to model the ability of intra-topic derivation. Specifically, we encode and restore the few training story text to serve as a reference to guide the generation at inference time.
Experimental results show that topic adaptation and prototype encoding structure mutually bring benefit to the few-shot model on BLEU and METEOR metric. The further case study shows that the stories generated after few-shot adaptation are more relative and expressive.
\end{abstract}



\maketitle

\section{Introduction}
\begin{figure}[tbp]
\centering 
\includegraphics[width=8.5cm]{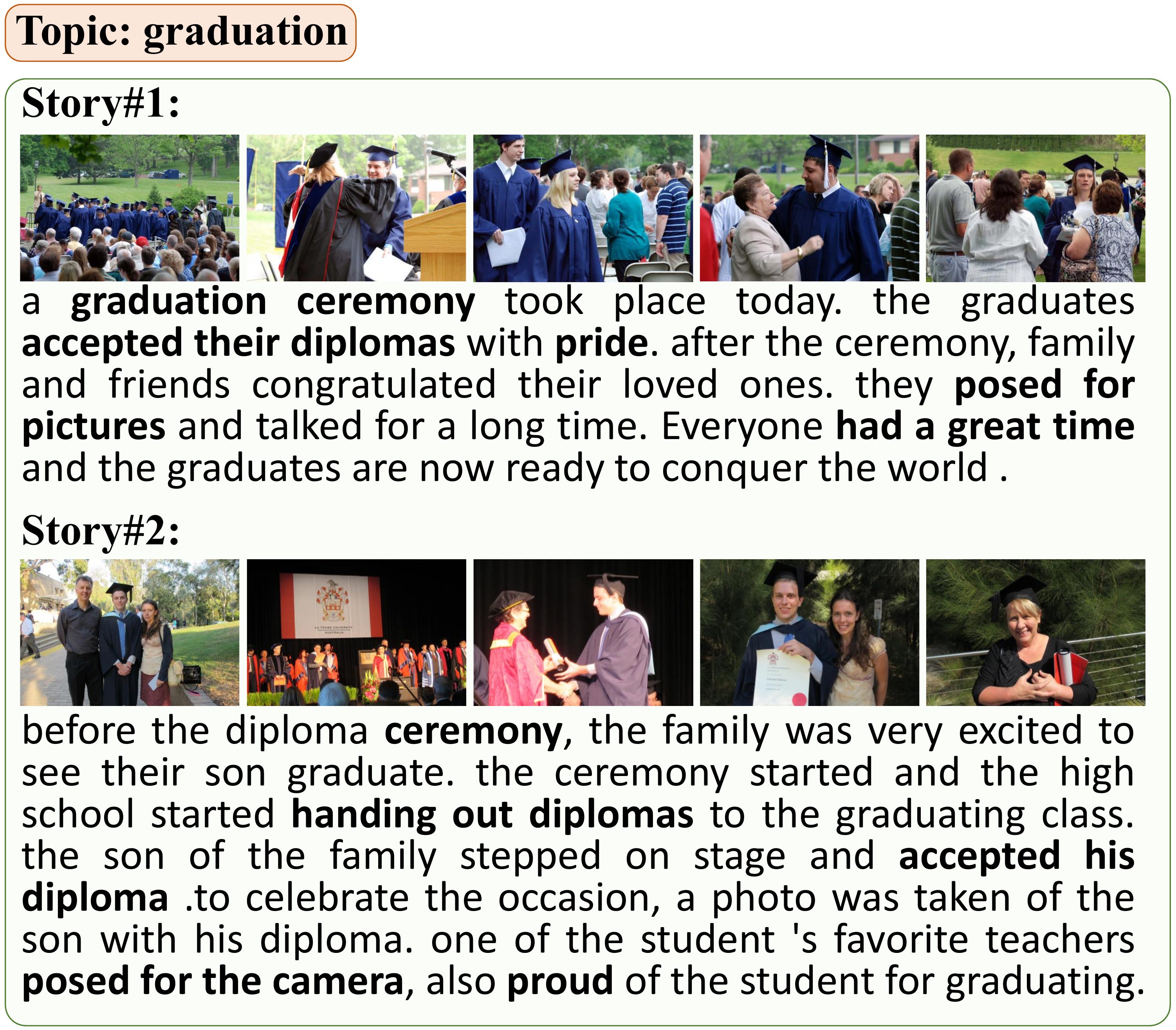} 
\caption{Story examples in the VIST dataset. The stories on the same topic have similar narrative styles such as sentence structure and word preference, as well as sentiment. The bold words are shared elements in the topic.} 
\label{figure1}
\end{figure}

Recently, increasing attention has been attracted to the Visual Storytelling~(VIST) task\cite{VIST}, which aims to generate a narrative paragraph for a photo stream, as shown in Figure \ref{figure1}. 
VIST is a challenge task because it requires deep understanding of spatial and temporal relations among images. The stories should describe the development of events instead of what is in the photos literally, adding complication to the generation. 

So far, many prior works on VIST~\cite{li2019informative,GLAC,HARNN,AREL,HSRL,WangHanQi17,vsimgdes,Consistent,Incorporating,knowledge,Hide-and-Tell,Character,Anchor} are focused on designing complex model structures. These models often require a huge amount of human-annotated data. However, the VIST annotating is costly and difficult, making it impractical to annotate a huge amount of new data. The workers are required to collect and arrange images about certain topics to form a photo stream, and then annotate each photo stream with a long narrative story, which should be well conceived, covering the idea of the topic. On the other hand, previous studies of topic model~\cite{LDA} suggest that real-world topics often follow the long-tail distribution. This means that in reality many new topics are not covered in the training dataset. The out-of-distribution~(OOD) has become a severe problem for applying a VIST model to the real-world scenario. To address the OOD problem, a VIST model should have the ability of topic adaptation. This requires that the VIST model can generalize to unseen topics after seeing only a few story examples. 

It is worth noticing that humans can quickly learn to tell a narrative story on a new topic given only a few examples. 1) Firstly, they have the abilities of inter-topic generalization. They can leverage the prior experience obtained from other related topics and integrate them with a few new examples to generalize from topic to topic. 2) Furthermore, they also have the abilities of intra-topic derivation. They can derive from the few stories on the same topic for their narrative style such as sentence structure and word preference, as well as the sentiment. We call these reference stories "prototype" in this paper. Take Figure~\ref{figure1} as an example, the story on graduation topic tends to describe shared elements such as "ceremony", "diploma", "proud" and "pose for pictures" etc. Both stories are narrated in positive sentiment. 

In this paper, we first propose the the few-shot visual storytelling task and further propose a Topic Adaptive Visual Storyteller~(TAVS) for this task. Our method is inspired by the humans way of telling a story. We model the abilities of both inter-topic generalization and intra-topic derivation mentioned above.

To model the ability of inter-topic generalization, we 
adopt the framework of Model-Agnostic Meta-Learning~(MAML)~\cite{MAML}, which is promising in few-shot tasks. We apply this framework to VIST so that the model can learn to generalize from topic to topic. We first train an initialization on sampled topics and optimize for the ability of quickly adapting to the new topic. Given a few story examples on the new topic, the model can adapt to the new topic in a few gradient steps at inference time. Both the visual encoder and the language decoder are adapted so that the storyteller can learn the ability to understand photo streams and organize stories simultaneously. 
Compared with standard VIST, TAVS optimize for the adaption ability instead of merely training an initialization parameter thus the adaptation is faster and easier.

To model the ability of intra-topic derivation, we further propose a prototype encoding structure for meta-learning on visual storytelling. It is intuitive to provide these annotations as the aforementioned "prototypes" for the generation model. Specifically, we encode and restore the few-shot examples at inference time into vector representations as "prototypes". The "prototypes" will be fused and go throughout the decoding stage to guide the generation. The model has access to not only vision but also text references, by which we are able to reduce the difficulty of few-shot learning.

It is noted that our prototype module is specially designed based on the few-shot setting and the proposed two parts are combined to form a whole. Compared with existing meta-learning methods~\cite{MAML,Retile}, this work focuses on the application of meta-learning on the vision-to-language generation problem. 
We explore a new way to exploit the topic information, i.e., thinking of "generating for a specific topic" as a meta-task and optimizing for the topic generalization ability.

Experiments on the VIST dataset demonstrate the effectiveness of our approach. Specifically, both meta-learning and prototype encoding bring benefits to the performance on BLEU and METEOR metrics. Besides, we rethink and discuss about the limitation of automatic metrics. To better evaluate our model, we also modify the "repetition rate" and "Ent score" proposed by the related task~\cite{planwrite,diversity} and introduce them to the VIST task. 
The further case study shows that the generated stories are more relative and expressive.

Our contributions can be summarized as follows:
\begin{itemize}
\item We propose a few-shot visual storytelling task which is closer to the real life situation compared with standard visual storytelling.
\item We further propose a topic adaptive storyteller combining the meta-learning framework and our own prototype encoding structure for the few-shot visual storytelling task. 
\item We evaluate our model on the VIST dataset and show two parts are mutually enhanced to benefit the performance in terms of BLEU and METEOR metric. The further case study shows that the generated stories are more relative and expressive.
\item We modify the "repetition rate" and "Ent-score" from the related works and introduce them to the VIST task for extra evaluation.

\end{itemize}
\begin{figure*}[htbp] 
\centering 
\includegraphics[width=17.8cm]{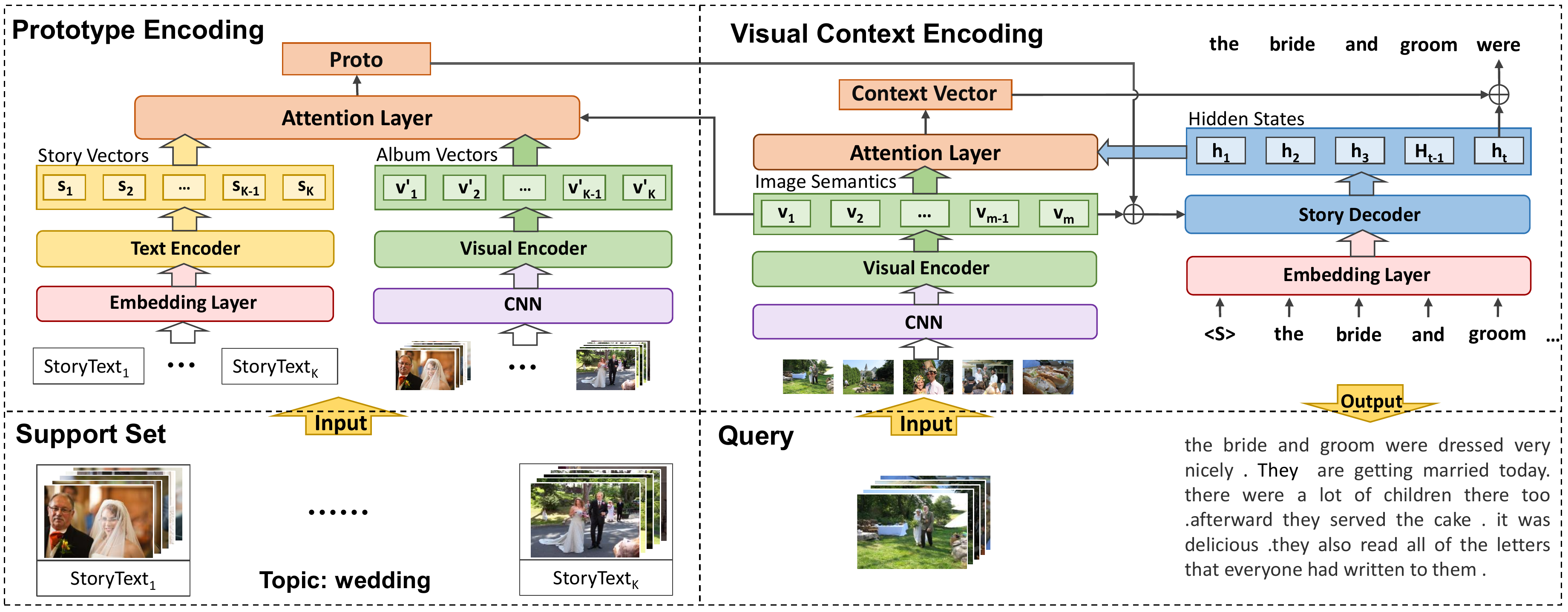} 
\caption{An overview of the generation model. We first extract story text features and album features of the support set. We use the vision information as the key for the attention to fuse the story text feature into prototype. The final hidden state of the visual encoder is concatenated with the fused prototype for decoder initialization. We also add a visual context vector by attention to relieve information loss during decoding.}
\label{figure3}
\end{figure*}
\section{Related Work}

\subsection{Visual Storytelling}
Prior works have built various seq2seq models since the task was proposed\cite{VIST}. These works have contributed to add complexity to model structures to improve the performance. 
\cite{HARNN,Sort} tries to sort images instead of using human arrangement. \cite{WangHanQi17} proposes a Contextual Attention Network to attend regions of photo streams for better visual understanding. 
\cite{aligment} tries to align vision and language in semantic space. 
Reinforcement Learning~(RL) is widely used in image captioning~\cite{required}. Later works also apply RL on the VIST task. 
\cite{AREL} points out the limitation of automatic metrics and further propose an adversarial reinforcement learning. 
\cite{HSRL} introduces hierarchical reinforcement learning algorithms to better organize the generation contents. 
\cite{Personal, Character} develops models to generate stories in different personality types. 
\cite{Post-Editing} proposes a post-editing strategy to boost performance. 
\cite{li2019informative} proposes cross-modal rules to discover abstract concepts from partial data and transfer them to the rest part for informative storytelling generation. \cite{knowledge2} using the knowledge graph to facilitate the generation. \cite{Hide-and-Tell} proposes an imagination network to learn non-local relations across photo streams. 
Prior works concentrate on the problems under the standard settings while this paper first considers this task in the few-shot scenario.
\subsection{Few-shot-learning for Seq2Seq Models}
Recently, meta-learning has gained its popularity in communities. It has shown great superiority in few-shot classification problems and is promising for seq2seq models. Algorithms such as MAML~\cite{MAML} and Reptile~\cite{Retile} are well adopted and the researchers have applied it to some language tasks. \cite{LRNMT,NLU} explores meta-Learning for low-resource neural machine translation and natural language understanding Tasks respectively. \cite{PDA} develops a personalizing dialogue agent via meta-learning. \cite{ZNNMT,TeaStu} focus on the low-resources or zero-shot neural translation tasks by transfer learning or distillation architecture. These works are focused on pure language tasks while this paper tackles a multi-modal task in meta-learning by considering the abilities of both visual understanding and language generation simultaneously. Though \cite{Fast} explores the few-shot Image captioning, they still convert the task into a classification problem while we explore the meta-learning for the cross-modal seq2seq structure. 

\subsection{Comparison with MAML}
The readers may be curious about the difference between the proposed TAVS and MAML since we have adopted the framework of MAML. (1)~First, MAML has shown its superiority in the area of regression, classification and reinforcement learning while we extend to to explore the effect of second-order optimization on a different challenge task, i.e., the vision-to-language problem.
(2)~Second,  although our TAVS consist of two parts, it is still a whole for the few-shot VIST task. Our prototype encoding module is designed based on the characteristic of few-shot language generation problem and itself can be viewed as an adaptation to the basic meta-learning model. The data point for the vision-to-language generation problem is in the form of "image-text" pair. Compared with a single category label in the classification problem, a text label contains rich information thus it can be further exploited during inference time. We adopt the framework as a part only to model the inter topic adaptation ability.  

\section{Approach}

\subsection{Problem Setting}
In this section, we first formalize the problem setting for the few-shot visual storytelling. Each story case consists of an input of 5 ordered images $X=\{I_1, ..., I_5\}$ and an output of a story text $Y=\{y_1, ..., y_n\}$. We define a task $\mathcal{T}$ in form of generating stories for photo streams on an identical topic $p(\mathcal{T})$. We view each topic as a distribution and the tasks are sampled from a specific topic, i.e., $\mathcal{T}{\sim}p(\mathcal{T})$.
We split topics with sufficient story cases to form the meta-training set $D_{train}=\{p_1(\mathcal{T}), ..., p_n(\mathcal{T})\}$ and those with few story cases to form the meta-testing set $D_{test}=\{p_1(\mathcal{T}), ..., p_m(\mathcal{T})\}$, where n and m are the number topics and n>m. 
For each topic $p(\mathcal{T})$, we sample few-shot cases to form the support set $D_{spt}$ for inner-training and some other cases to form query set $D_{qry}$ for inner-testing, i.e., $\mathcal{T}=\{D_{spt}, D_{qry}\}$. Thus for the K-shot VIST task, each task sample is comprised of 2K story cases sampled from an identical topic; K for $D_{spt}$ and another K for $D_{qry}$, i.e., $|D_{spt}|=|D_{qry}|=\text{K}$. 
Then, we can train a Topic--Adaptive-Visual-Storyteller(TAVS) model to learn from commonly seen topics and adapt quickly to new topics with fewer story samples. In effect, the model gets new parameters on $D_{spt}$ and its performance is further estimated on $D_{qry}$ to show how well does the model learns from K story cases.

\subsection{Base Model}
Figure~\ref{figure3} shows an overview of our TAVS model. For computational efficiency, we extract all image features on VIST datatset with Resnet152 during pre-processing. The input of the model is 5 image features of 2048-dimension vectors in fact. These features are encoded by a \emph{Visual Encoder} for visual context understanding:
\begin{equation}
    V’ = \{v_1,...,v_5\} = \text{GRU}_{VE}(I_1,..,I_5)
\end{equation}
Then we initialize the story decoder with the last hidden state $v_5$ of the \emph{Visual Encoder}. The initial state is then decoded to a long story text. Both \emph{Visual Encoder} and \emph{Story Decoder} are implemented by a single layer GRU for saving memory. The decoding process can be formalized as:
\begin{align}
    \label{eq2}h_0 &= W_{init}\cdot{v_5} \\
    \label{eq3}h_t &= \text{GRU}_{SD}(h_{t-1},E\cdot{y_{t-1}}) \\ 
    \label{eq4}p_{y_t} &= \text{softmax}(\text{MLP}(h_t))
\end{align}
where $W_{init}$ is a learned projection, $E$ is the learned word embedding matrix, $GRU_{SD}$ refer to the specific GRU for story decoder, MLP($\cdot$) represents the multi-layer perception, $p_{y_t}$ is the word probability distribution.
To map from an image sequence to a word sequence, the model parameter is optimized by standard Max Likelihood Estimation~(MLE):
\begin{equation}
    \label{eq5}\mathcal{L}(\theta)=\sum_{l=1}^{n}
    -\log{p(y_l|y_1,...y_{l-1},v_1,...,v_5,\theta)}
\end{equation}
In practice, the loss function is calculated by the cross-entropy between predicted words and ground truths. 

It is noted that prior works~\cite{AREL,li2019informative,HSRL,GLAC,HARNN} decode 5 hidden states separately while we design the model in this way since we will add the prototype information later. So we treat 5 annotated sentences as a whole to see the effect of prototype information.

\subsection{Visual Context Encoding}
Encoding all visual features into a single hidden state may lead to information loss. Besides, the information in the hidden state diminishes with the growth of the decoding step. These factors result in repeated and boring stories during generation. To tackle this problem, we add a visual attention mechanism to retain visual information during decoding. Different from spatial attention which attends to regions of an image, the model attend to the hidden state $\mathcal{V}$ of \emph{Visual Encoder}. We adopt the attention in \cite{Attention}:
\begin{equation}
    \label{att}
    \textbf{Attention}(Q,K,V)=\text{softmax}(\frac{QK^T}{\sqrt{d_k}})V
\end{equation}
where $d_k$ is the dimension of $K$.
We adopt it since it is efficient without extra parameters. The computation of parameter update is more costly for second-order optimization. Formally, we set the current hidden state $h_t$ of the \emph{Story Decoder} as Query $Q$ and hidden states $\mathcal{V}$ of the \emph{Visual Encoder} as Key $K$ and Value $V$. The final visual context vector is calculated by:
\begin{equation}
    c_t = \textbf{Attention}(h_t,\mathcal{V},\mathcal{V})
\end{equation}    
  Then the visual context vector is concatenated with decoder hidden state at decoding step $t$ for word prediction.
 Now we can rewrite the equation \ref{eq4} as:
 \begin{equation}
    p_{y_t} = \text{softmax}(\text{MLP}([h_t;c_t]))
\end{equation} 
where [;] represents concatenating.

\subsection{Prototype Encoding}
\label{sec:3.4}
For few-shot language generation problems, the long text annotation can be encoded to provide a basic reference which we call "prototypes" in this paper. As for visual storytelling, the story text of few-shot examples for inner-training are encoded by a text encoder and restored for retrieval during inner-testing. 
First, the story texts in $D_{spt}$ are encoded as story text vectors:
\begin{align}
    &s_i=\textbf{TextEncoder}(Y_i), Y_i\in{\mathcal{D}_{spt}} \\
    &\mathcal{S}_{spt}=\{s_1, ..., s_n\}
\end{align}
where TextEncoder is implemented by another GRU called $GRU_{TE}$, $Y_i$ is the story text in the support set $\mathcal{D}_{spt}$, $\mathcal{S}_{spt}$ is the set of story text vectors.
Then we use the attention mechanism in Equation~(\ref{att}) as well to fuse the stories in the support set. We represent a photo stream by the final hidden state $v_5$ of the Visual Encoder. By setting the testing photo stream as Query $Q$, few-shot training photo streams $\mathcal{V}_{spt}$ as Key $K$ and the encoded story vectors $\mathcal{S}$ as Value $V$, we get a prototype vector which is a weighted sum of story vectors:
\begin{equation}
    proto=\textbf{Attention}(v_{qry},\mathcal{V}_{spt},\mathcal{S}_{spt})
\end{equation}
For the situation where we only deal with the support set(e.g., the inner update of meta-learning, the supervised model for ablation study ), we still allow the model to attend to exmaples in the supprot set:
\begin{equation}
    proto=\textbf{Attention}(v_{spt},\mathcal{V}_{spt},\mathcal{S}_{spt})
\end{equation}
In effect, such calculation with attention weights is equivalent to a measure of similarity. After retrieving the prototype vector, it is sent together with the visual hidden to a Multi-Layer Perceptron~(MLP) for decoder initialization. We rewrite the equation \ref{eq2} as:
\begin{equation}
    h_0=W_{init}\cdot{[v_{qry};proto]}
\end{equation}
In this way, the prototype will guide the generation throughout the decoding step.
Then the loss function of equation~(\ref{eq5}) for single story example can be written as:
\begin{equation}
    \mathcal{L}(\theta)=
    -\log({Y_{qry}|X_{qry}, \mathcal{S}_{spt}, \mathcal{V}_{spt}, \theta)}
\end{equation}
where $X_{qry}$ is a single photo stream in the query set, $Y_{qry}$ is the corresponding target story, $\mathcal{V}_{spt}$ is all the photo stream feature in the support set, $\mathcal{S}_{spt}$ are the corresponding story text features.

\begin{figure}[tbp]
\centering 
\includegraphics[width=8.5cm]{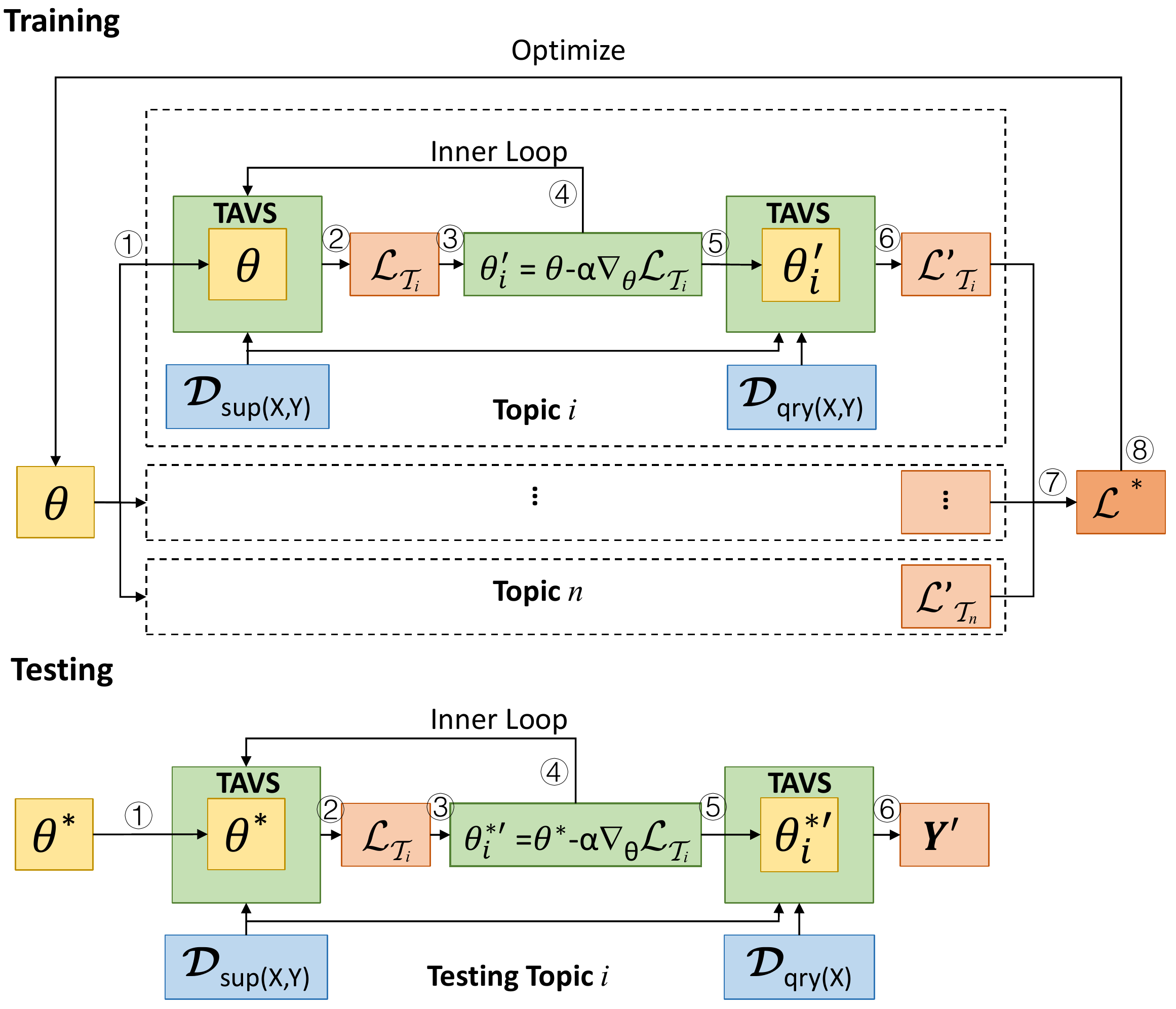} 
\caption{The meta-training and the meta-testing process of TAVS. The initial parameter is adapted to a new parameter for a specific topic. We optimize the initial parameter by the meta loss of training topics. The support set will be reused to provide prototypes at inference time.} 
\label{figure4}
\end{figure}

\subsection{Training Storyteller with Meta-Learning}
\label{sec:3.5}
\subsubsection{Gradient-Based Meta-Learning}
Gradient-Based Meta-learning algorithms optimize for the ability to learn new tasks quickly and efficiently.
They learn from a range of meta-training tasks and are evaluated based on their ability to learn new meta-test tasks. Generally, meta-learning aims to discover the implicit structure between tasks such that, when the model meets a new task from the meta-test set, it can exploit such structure to quickly learn the task. 

\subsubsection{Meta Training for Visual Storyteller}
The core idea of meta-learning for VIST is to view visual storytelling for a specific topic as a task and optimize for the ability to adapt to a new topic. As shown in Figure~\ref{figure4}, we specifically illustrate the meta-training and meta-testing process for VIST. Suppose we have built a base model $f_\theta$ that can map from photo stream $X$ to story text $Y^{'}$, for each topic, the initial data parameter $\theta$ is updated for M steps:
\begin{equation}
    \theta_{i}^{'}\leftarrow \theta-\alpha\nabla_{\theta}\mathcal{L}_{\mathcal{T}_i}
\end{equation}
where the loss $\mathcal{L}_{\mathcal{T}_i}$ is calculated on the support set $\mathcal{D}_{sup}$ by equation(\ref{eq5}), $\alpha$ is the inner update learning rate. The parameter $\theta_{i}^{'}$ is further tested on the query set $\mathcal{D}_{qry}$ to get meta loss $\mathcal{L}_{\mathcal{T}_i}^{'}$. At each training iteration, we sample a batch of topics from $p\mathcal{(T)}$ and optimize the initial data parameter as follows:
\begin{equation}
    \min_{\theta}\mathbb{E}\left[\mathcal{L}_{\mathcal{T}_i}(f_{\theta_{i}^{'}})\right]
    =\frac{1}{N}\sum_{\mathcal{T}_{i}{\sim}p(\mathcal{T})}\mathcal{L}_{\mathcal{T}_i}
    (f_{\theta-\alpha\nabla_{\theta}\mathcal{L}_{\mathcal{T}}({f_\theta})})
\end{equation}
where $N$ is the number of topics in a batch. More concretely, the optimization is first performed by Stochastic Gradient Descent~(SGD):
\begin{equation}
    \theta\leftarrow\theta-\beta\frac{1}{N}\nabla_{\theta}
    \sum_{\mathcal{T}_{i}{\sim}p(\mathcal{T})}\mathcal{L}_{\mathcal{T}_i}(f_{\theta_{i}^{'}})
    \label{eq15}
\end{equation}
where $\beta$ is the meta step size. In our experiments, we find that other adaptive optimizers such as Adam is also compatible and we use it for faster training. 

We optimize for an initial parameter setting such that one or a few steps of gradient descent on a few data points lead to effective generalization. Then, after meta-training, we follow the Equation~(\ref{eq15}) to fine-tune the learned parameters to adapt to a new topic in meta-testing. This meta-learning process can be formalized as learning a prior over functions, and the fine-tuning process as inference under the learned prior.

It is also noted that a seq2seq model has different settings during training and testing. Alternatively, we can input the ground truth word~(Teacher Forcing) or the last word predicted by the model~(Greedy Search). For training stability, we use teacher forcing for both the inner update and the meta update during meta-training. During meta-testing, we still use teacher forcing to update parameters on few-shot examples, and beam search is used during the final generation stage. The beam search algorithm keeps candidate sequences with top $K$ probability.


\section{Experiments and Analysis}

\subsection{Experimental Setup}

\subsubsection{VIST Dataset}
VIST(SIND v.2) is a dataset for sequential-vision-to-language, including 10,117 Flickr albums with 81,743 unique photos in 20,211 sequences, aligned to descriptive and story language. It is previously known as "SIND", the Sequential Image Narrative Dataset (SIND). The samples in VIST are now organized in a story form which contains a fixed number of 5 ordered images and aligned story language. The images are selected, arranged and annotated by AMT workers and one album is paired with 5 different stories as reference. There are 50,200 story samples in total annotated with 69 topics. We first filter out stories with broken images and too long or too short text, leaving 43838 qualified stories. Then, we sort the topics by the number of stories in them. The top 50 frequent topics are used for meta-training and the remainders are for meta-testing. 

\begin{table*}[t]
\centering
\caption{Zero-shot and few-shot results on the VIST dataset. $\uparrow$ indicates higher better while $\downarrow$ indicates lower better. The proposed TAVS has a significant improvement on all evaluations, showing the effect of topic adaptation and prototype encoding.}
\begin{tabular}{l|cc|ccc|ccc}  
\toprule
\multicolumn{1}{c|}{\multirow{2}{*}{Model}} & \multicolumn{2}{c|}{Settings} & \multicolumn{3}{c|}{5-Shot}   & \multicolumn{3}{c}{Zero-SHot}    \\ \cline{2-9} 
\multicolumn{1}{c|}{}                       & \multicolumn{1}{c}{inter(ML)} & \multicolumn{1}{c|}{intra(PE)} & \multicolumn{1}{c}{BLEU$\uparrow$} & \multicolumn{1}{c}{METEOR$\uparrow$} & \multicolumn{1}{c|}{NLL$\downarrow$} & \multicolumn{1}{c}{BLEU$\uparrow$} & \multicolumn{1}{c}{METEOR$\uparrow$} & \multicolumn{1}{c}{NLL$\downarrow$} \\
\hline
Supervised(baseline) &$\times$   &$\times$   &6.3          &29.0          &6.125          &5.7          &28.8          &13.480\\
Proto-Supervised     &$\times$   &\checkmark &6.4          &29.5          &5.772          &6.2          &29.1          &13.045\\
Meta-Learning        &\checkmark &$\times$   &8.1          &30.5          &5.630          &7.1          &30.0          &11.856\\
TAVS(full model)     &\checkmark &\checkmark &\textbf{8.1} &\textbf{31.1} &\textbf{5.532} &\textbf{7.3} &\textbf{30.7} &\textbf{11.285} \\

\bottomrule
\end{tabular}
\label{ablation}
\end{table*}

\subsubsection{Evaluation}
The evaluation of VIST is commonly done by metrics such as BLEU and METEOR. We utilize the open-source evaluation code provided by Yu et al.~\cite{HARNN}. 
The BLEU score indicates the reappearance rate of short phrases. METEOR is a more complex metric which takes synonyms, root and order into consideration. We set METEOR score as the main indicator since Huang et al.~\cite{VIST} concludes that METEOR correlates best with human judgment according to correlation coefficients. 
Regarding the CIDEr metric, it is pointed out by \cite{AREL} that ”CIDEr measures the similarity of a sentence to the majority of the references. However, the references to the same image sequence are photostream different from each other, so the score is very low and not suitable for this task.” The situation even gets worse under the few-shot setting thus we do not use this metric.

The existing metrics only indicate the overlapping level of the generated stories and human-annotated stories. They are not sufficient to show the topic adaptation abilities of models, thus we further design a new indicator for adaptation ability evaluation. We first train a topic classifier to predict the topic of the generated stories and use the posterior of the topic classifier as our indicator. The posterior is represented by the negative log likelihood~(NLL) of the true topic. In effect, it is equivalent to the cross-entropy loss of the multi-classifier. A smaller NLL indicates the generated stories are more related to the true topic.

\subsubsection{Training Details}
All of the encoders and decoders in our model are implemented by a single layer GRU with hidden size 512. For each iteration, we sample 5 topics and update 3 times to save memory. We observe a decrease in the average inner training loss with more update times but no explicit performance gain in inner testing. The inner update is implemented by SGD with a learning rate of 0.05 and the meta optimization is implemented by Adam with a learning rate of 0.001. We also add a dropout of 0.2 and a gradient clipping of 10. It is pointed out that we do not update the word embedding at inner update because the word embedding matrix will go to a stable state during training. This kind of universal word embedding contributes a lot to the saving of memory and computation. Besides, we move out repeated sentences in post-processing to reduce redundancy.

\subsection{Ablation Study}
In order to demonstrate the effectiveness of our model, we carried out experiments on the VIST dataset. Our settings are:
\begin{itemize}
    \item \textbf{Supervised} A supervised model on meta-training set. It is noted that the data in a batch comes from different topics and there is no inner update. 
    \item \textbf{TAVS} A model trained by meta-learning with prototype encoding. This is our full model. We model the abilities of both inter-topic generalization and intra-topic derivation.
    \item \textbf{Proto-Supervised} A supervised model with prototype encoding described in section~\ref{sec:3.4}. Only intra-topic derivation ability is modeled. We set this model to see the effect of prototype encoding.
    \item \textbf{Meta-Learning} A model trained by meta-learning as described in section~\ref{sec:3.5}. The data are fed on task level. Only inter-topic generalization ability is modeled. We set this model to see the effect of meta-learning.
\end{itemize}
\subsubsection{Effect of Joint Model}
Table~\ref{ablation} shows the performance of different settings.
Compare the "Supervised" model with the "TAVS" model, we can find a significant improvement on all evaluations, showing the effectiveness of our approach. There is a relative improvement of 28.6\% and 7.2\% for BLEU and METEOR on few-shot learning, 28.1\% and 6.6\% for zero-shot learning, indicating the stories generated by "TAVS" have better quality. The "TAVS" model also gets a lower NLL, indicating the stories generated by it have a better topic correlation with the true topic. 

\subsubsection{Effect of Prototype Encoding}
Compare the "Supervised" model with the "Proto-Supervised" model, we can find that the model with prototype encoding has a higher METEOR score, lower NLL and similar BLEU score in the 5-shot scenario. The improvement for METEOR is 1.7\%. We can infer that the prototype encoding mainly contribute to the improvement of NLL and METEOR score and has less effect on BLEU score.
Compare the "Meta-Learning" model with the "TAVS" model, we get the same conclusion. Even Meta-Learning has raised the METEOR score, the Prototype Encoding can further improve the score for 2.0\%.

\subsubsection{Effect of Meta-Learning}
Compare the "Supervised" model with the "Meta-Learning" model, we can find that the model trained by meta-learning has an overall improvement in the 5-shot scenario. The improvement for BLEU and METEOR is 28.6\% and 5.2\% respectively.
Compare "Proto-Supervised" model with the "TAVS" model, the improvement for BLEU and METEOR is 26.6\% and 4.5\% respectively. The result is consistent with the last comparison.

\subsubsection{Effect of Few-Shot Learning}
Compare the result of "Zero-Shot" and "5-shot", we can find that the model adapting after 5 examples performs better. 
It is also noted that "Proto-TAVS" models already get relatively high scores in zero-shot setting. This means the initialization parameter trained by meta-learning has obtained better language ability before adaptation. 
We speculate that topic-batched data and second derivatives contribute to the result. 

\begin{table}[t]
\centering
\caption{Comparison with SOTA methods. Our approach gets the best result on the metric evaluation.}
\setlength{\tabcolsep}{2.5mm}{
\begin{tabular}{c|ccc}  
\toprule
Model &{\quad}BLEU4$\uparrow$ & METEOR$\uparrow$ &NLL$\downarrow$\\
\hline
XE~\cite{AREL} &7.4 & 30.2 &5.743\\
GLAC~\cite{GLAC} &7.8 &30.1 &5.762\\
RL-M~\cite{Rennie} &6.6 &30.5 & 5.646\\
RL-B~\cite{Rennie} &7.6 &30.0 &5.751\\
AREL~\cite{AREL} &7.6 &30.6 &5.613\\
\hline
Supervised(ours) &6.3 &29.0 &6.125\\
TAVS(ours) &\textbf{8.1} &\textbf{31.1} &\textbf{5.532}\\
\bottomrule
\end{tabular}
}
\label{cmp}
\end{table}
\subsection{Comparison with SOTA}
In order to show the effectiveness of our methods, we also try to compare with some SOTA methods. However, the scores of these methods are reported under the standard setting while we first explore the few-shot setting for this task, thus we cannot directly copy the scores and compare with them. So we adapt the few typical works to fit the few-shot setting for comparison. It is noted that, though \cite{Whatmake,knowledge} get higher score under the standard setting, we do not compare with them since they have used extra resources such as "Pretrained BERT" and "Knowledge Graph". The descriptions for these models are as follows:
\begin{itemize}
    \item \textbf{XE} A supervised model trained by cross entropy. The model extracts 5 context features and decodes each of them into a sentence with a shared decoder. The model generates a fix number of sentences and concatenate them to form a story.
    \item \textbf{GLAC} A supervised model with "GLocal attention". The stories are also obtained through concatenation.
    \item \textbf{RL-M} A model trained by reinforcement learning with METEOR scores as reward. 
    \item \textbf{RL-B} A model trained by reinforcement learning with BLEU scores as reward.
    \item \textbf{AREL} A model trained by Adversarial Reward Learning. The reward is provided by a discriminator network which is trained alternately with the generator.

\end{itemize}

The scores are reported in the Table~\ref{cmp}. Our TAVS gets the highest score on both BLEU and METEOR. It is noted that our model does not rely on any explicit meanings of the topic. The topic information is implicitly contained in the data, which is caused by splitting and sampling. Therefore, the superiority of our model relies on better adaptation ability, instead of extra information.

We are also impressed by the high score achieved by the SOTA methods since they are originally designed for large amounts of data. However, we find that the stories generated by these models tend to be general which is not reflected by the metrics scores. Thus we try to evaluate the models by other measurements.

\subsection{Rethinking the Limitation of Metrics}
Wang et al.~\cite{AREL} has pointed out the limitation of metrics such as BLEU and METEOR. The metrics are based on the overlap between the generated story and the ground truth. Thus a relevant story may get a lower score while a meaningless story with only preposition and pronoun has the chance to cheat scores. Compared with image captions, stories are longer and more flexible, even with some imagination. Thus in our view, it is more like a stylized generation task than a pure visual grounding task. So we try to set up other evaluations such as diversity to verify our model. 

\begin{table}[t]
\centering
\caption{The inter \& intra repetition rate and Ent score of the models. The TAVS gets lower repetition rate and higher Ent score.}
\setlength{\tabcolsep}{1.2mm}{
\begin{tabular}{c|ccc|ccc}  
\toprule
\multicolumn{1}{c|}{\multirow{2}{*}{Models}} &\multicolumn{3}{c|}{4-gram} &\multicolumn{3}{c}{5-gram}\\
\cline{2-6}
\multicolumn{1}{c|}{} &inter$\downarrow$ &intra$\downarrow$ &Ent$\uparrow$ &inter$\downarrow$ &intra$\downarrow$ & Ent$\uparrow$\\
\hline
XE~\cite{AREL}    &89.6\%          &4.6\%          &2.947          &83.9\%          &2.5\%      &2.869\\
GLAC~\cite{GLAC}  &86.0\%          &4.0\%          &2.682          &78.7\%          &2.3\%      &2.450\\ 
RLB~\cite{Rennie}  &91.8\%         &7.6\%          &2.814          &86.4\%          &4.8\%      &2.648\\
RLM~\cite{Rennie}  &89.2\%         &5.5\%          &2.894          &83.2\%          &3.7\%      &2.671\\
AREL~\cite{AREL}  &86.8\%          &3.9\%          &3.088          &79.6\%          &2.4\%      &2.910\\
\hline
Supervised        &90.8\%          &3.5\%          &2.544          &88.2\%          &1.9\%     &2.561\\
TAVS              &\textbf{75.7\%} &\textbf{1.8\%} &\textbf{3.146} &\textbf{69.5\%} &\textbf{1.2\%} &\textbf{3.071}\\
\bottomrule
\end{tabular}
}
\label{rep}
\end{table}
\subsubsection{Repetition Rate}
In order to further evaluate the diversity of the models, we calculate the repetition rate of the generated stories. ~\cite{planwrite} proposes two measurements to gauge inter- and intra-story repetition for the text generation task. We are inspired by there work and further extends the original trigram setting to n-gram setting to evaluate our models. The inter-story repetition rate $r_e$ and intra-story repetition rate $r_a$ are computed as follows:
\begin{align}
    r_e &= 1-\frac{|\sum_{i=1}^N\sum_{j=1}^{m}f_n^*(s_{i,j})|}
    {\sum_{i=1}^N\sum_{j=1}^{m}|f_n(s_{i,j})|} \\
    r_a^j &= \frac{1}{N}\sum_{i=1}^{N} \frac{\sum_{k=1}^{j-1}|f_n(s_{i,j})\cap{f_n}(s_{i,k})|}
    {(j-1)*|f_n(s_{i,j})|} \\
    r_a &= \frac{1}{m}\sum_{j=1}^{m}r_a^j
\end{align}
where $s_{i,j}$ represents the j-th sentence of the i-th story, $f_n$ return the set of n-grams of the sentence, $f_n^*$ return the distinct set of n-grams, $|\cdot|$ means the number of element of a set. 

In effect, the inter-story repetition rate computes the proportion of repeated n-grams in all n-grams. The intra-story repetition rate first computes the repeat proportion of the j-th sentence compared with previous sentences and then average them.

\subsubsection{Entropy Metric}
We also find a similar idea of "Dist-n" in~\cite{diversity} which computes the proportion of distinct n-grams in all n-grams. The measure is equivalent to the inter-repetition-rate. The only difference is that it does not need to be subtracted by 1.
However, \cite{entn} pointed out that these kinds of metrics neglect the frequency difference of n-grams. There is a example from \cite{entn}: token A and token B that both occur 50 times have the same Dist-1 score (0.02) as token A occurs 1 time and token B occurs 99 times, whereas commonly the former is considered more diverse that the latter. Thus it proposed "Entropy (Ent-n)" metric, which reflects how evenly the empirical n-gram distribution is for a given sentence:
\begin{equation}
    Ent(n) = -\frac{1}{\sum_{w_n}{F(w_n)}}
    \sum_{w_n\in{V}}{F(w_n)log\frac{F(w_n)}{\sum_{w_n}{F(w_n)}}}
\end{equation}
where V is the set of all n-grams, F($w_n$) denotes the frequency of n-gram w.

Ent can also be used to evaluate the VIST task. For a long story text, we view it as a sentence and remove the n-grams containing a punctuation since they have crossed the sentences.

\subsubsection{Analysis}
The "repetition rate" and the "Ent score" of each model are shown in the Table~\ref{rep}. The previous works have a higher inter \& intra repetition and a lower Ent score in the few-shot setting, though they get high metric scores. They are more likely to generate general sentences which are suitable for all topics. Meanwhile, our TAVS achieves the lowest inter and intra repetition rate and higher Ent score, showing the effect of topic adaptation. 
\begin{figure*}[t] 
\centering 
\includegraphics[width=17.8cm]{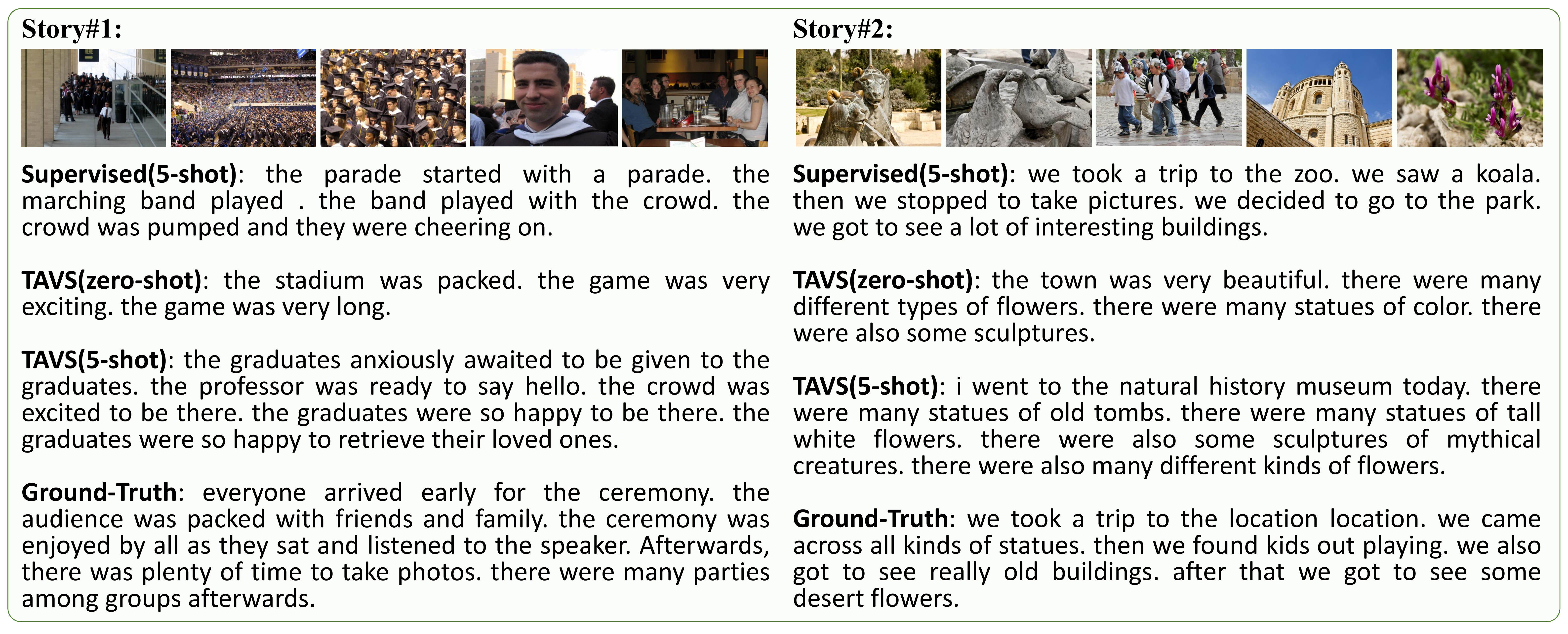} 
\caption{Examples of stories generated by different models. The stories generated after few-shot adaptation are more relative and expressive.}
\label{figure6}
\end{figure*}

\begin{table}[t]
\centering
\caption{The effect of updated learning rate.}
\setlength{\tabcolsep}{2mm}{
\begin{tabular}{c|cc|cc}  
\toprule
\quad &\multicolumn{2}{c}{5-Shot} &\multicolumn{2}{c}{Zero-shot} \\
\hline
ULR &BLEU4 &METEOR &BLEU4 &METEOR \\
\hline
0.01 &7.7 &30.9 &7.2 &30.5 \\
0.03 &8.1 &30.9 &\textbf{7.8} &30.7 \\
0.05 &\textbf{8.1} &\textbf{31.1} &7.3 &\textbf{30.7} \\
0.07 &7.5 &30.7 &7.0 &30.1 \\
0.1 &7.6 &30.5 &0.01 & 5.9\\
\bottomrule
\end{tabular}
}
\label{LR}
\end{table}
\begin{figure}[t]
\centering 
\includegraphics[width=8.5cm]{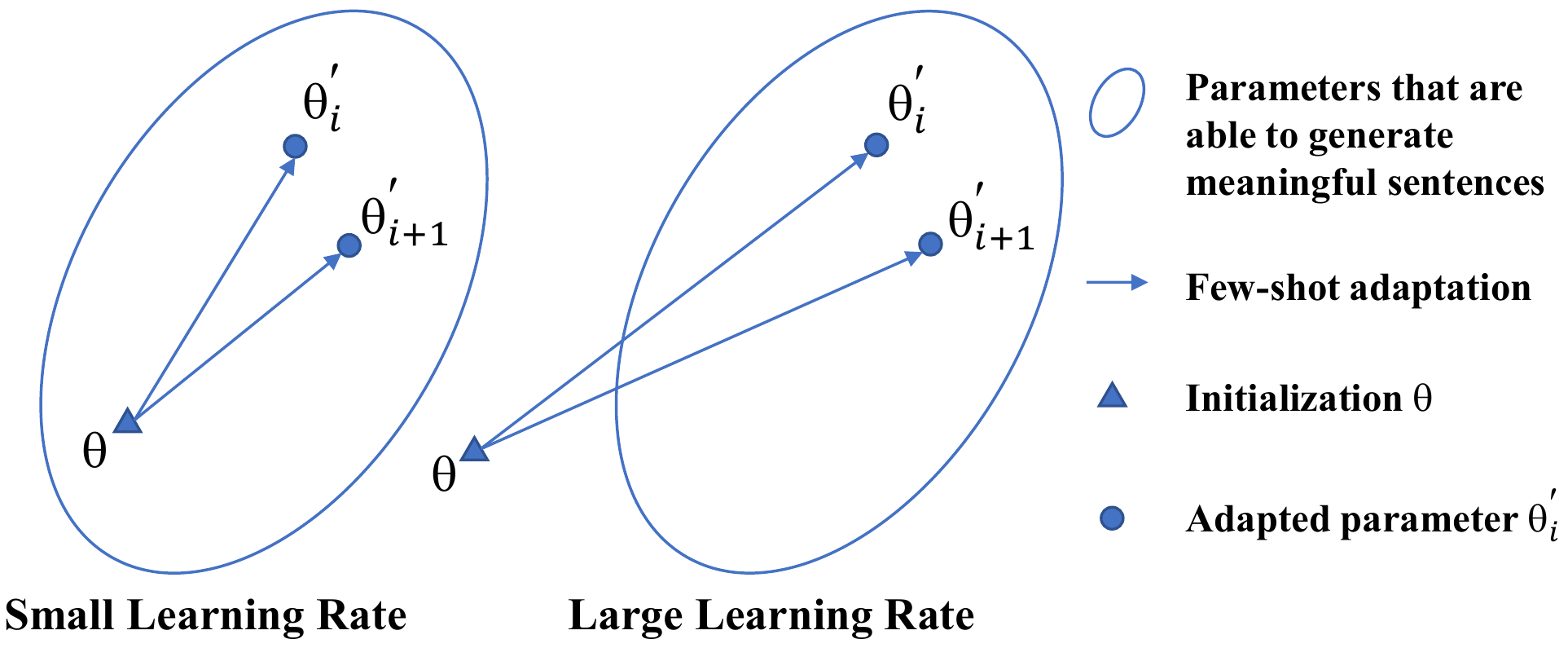} 
\caption{Comparison of different update learning rate. When ULR is larger, the initial parameter point is out of the range where the model is able to generate meaningful sentences. However, the model finds a trick to use gradients of few-shot examples as a springboard to return to a reasonable parameter point. } 
\label{figlr}
\end{figure}
\subsection{Effect of Update Learning Rate}
During training, we find that the inner update learning rate~(ULR) is an important hyperparameter for the model. It determines how much the model learns from the few-shot examples when adapting. 

The effect of ULR is shown in Table~\ref{LR}. We tune the ULR from 0.01 to 0.05, the metric scores get higher and the generated stories get more changes. We further tune the ULR to 0.07, the scores do not increase any more. When the ULR is 0.1, the zero-shot scores are seriously damaged while the 5-shot scores are still at a normal level. Then we further investigate into such an abnormal phenomenon. 

When the ULR is 0.1, we observe that the model gets a high inner loss but a reasonable meta loss. In this case, the stories generated by the zero-shot model are totally rubbish while those generated by the few-shot model are meaningful. As illustrated in Figure~\ref{figlr}, this indicates that When ULR is larger, the initial parameter point is out of the range where the model is able to generate meaningful sentences. However, the model finds a trick to use gradients of few-shot examples as a springboard to return to a reasonable parameter point. Under this circumstance, the model can not work in the zero-shot setting but still works well in the few-shot setting. We conclude that the model should learn properly from the examples and it has a high tolerance. When we tune the ULR higher to 0.2, the model do not converge. This indicates the limitation of the tolerance.


\subsection{Case Study}
Figure~\ref{figure6} shows examples of stories generated by different models. Compare the stories generated by the "Proto-TAVS" model and the "Supervised" model, the "Supervised" model struggles to tell a story on the true topic while "Proto-TAVS" model can fit the true topic successfully. We also compare the stories generated by the zero-shot model and few-shot model to show the effect of adaptation. In story1, the zero-shot model generates a story about "game" mistakenly while it generates a story about "graduation" correctly after the few-shot adaptation. It is noted that the content is more coherent and concrete. Story2 shows another case that the story generated by the zero-shot model is coherent but the one generated by the few-shot model is more relative and expressive. This indicates the model has the ability to learn from few-shot examples and adapt to new topics. We find that the zero-shot model already has certain abilities of visual understanding and language generation, which is different from the classification scenario.

\section{Conclusion}
In this paper, we explore the meta-learning for visual storytelling and further propose a prototype encoding architecture to enhance the meta-learning. Our model can learn from few-shot examples and adapt to a new topic quickly with better coherence and expressiveness. We believe there is still much space for improvement. Many problems such as better diversity, effective evaluation and relation modeling need more exploration. 

\section*{Acknowledgments}
This work has been supported in part by National Key Research and Development Program of China (2018AAA010010), NSFC (U1611461, 61751209, U19B2043, 61976185), Zhejiang Natural Science Foundation (LR19F020002, LZ17F020001), University-Tongdun Technology Joint Laboratory of Artificial Intelligence,Zhejiang University iFLYTEK Joint Research Center, Chinese Knowledge Center of Engineering Science and Technology (CKCEST), the Fundamental Research Funds for the Central Universities, Hikvision-Zhejiang University Joint Research Center.

\bibliographystyle{ACM-Reference-Format}
\bibliography{mmbib}


\end{document}